\pdfoutput=1

\documentclass[11pt]{article}
\usepackage{lmodern}
\usepackage{EMNLP2023}
 
\usepackage{graphicx}
\usepackage{amsmath}
\usepackage{amssymb}
\usepackage{booktabs}
\usepackage{multirow}
\usepackage{enumitem}
\usepackage[title]{appendix}
\usepackage{makecell}
\usepackage{amsmath}
\usepackage{amssymb}
\usepackage{bm}
\usepackage{xcolor}
\usepackage{booktabs} 
\usepackage{multirow} 
\usepackage{colortbl}
\usepackage{booktabs}
\usepackage{multirow}
\usepackage{adjustbox}
\usepackage{subcaption}
\usepackage[T1]{fontenc}
\usepackage{times}

\usepackage[utf8]{inputenc}

\usepackage{microtype}

\usepackage{inconsolata}

\title{Spotlighter: Revisiting Prompt Tuning from a Representative Mining View}

\author{
  Yutong Gao\textsuperscript{1*}, 
  Maoyuan Shao\textsuperscript{1*}, 
  Xinyang Huang\textsuperscript{2},
  Chuang Zhu\textsuperscript{2},\\
\textbf{Lijuan Sun} \textsuperscript{3†},
  \textbf{Yu Weng}\textsuperscript{1},
  \textbf{Xuan Liu} \textsuperscript{1†},
  \textbf{Guoshun Nan}\textsuperscript{4}\\
  \textsuperscript{1}School of Information Engineering, Minzu University of China \\
  \textsuperscript{2}School of Artificial Intelligence, Beijing University of Posts and Telecommunications\\
  \textsuperscript{3} National Library of China, Beijing, China\\
  \textsuperscript{4} School of Cyberspace Security, Beijing University of Posts and Telecommunications\\
  \texttt{\small{\{ytgao92,maoyuanshao,wengyu,liuxuan\}@muc.edu.cn, \{chuangzhu,xinyanghuang,sunlijuan\}@bupt.edu.cn}}
}

\begin{document}
\maketitle

\renewcommand\thefootnote{} 
\footnotetext{* These authors contributed \textbf{equally} to this work.}
\footnotetext{† Corresponding authors.}
\begin{abstract}
CLIP’s success has demonstrated that prompt tuning can achieve robust cross-modal semantic alignment for tasks ranging from open-domain recognition to fine-grained classification. However, redundant or weakly relevant feature components introduce noise and incur unnecessary computational costs. In this work, we propose Spotlighter, a lightweight token-selection framework that simultaneously enhances accuracy and efficiency in prompt tuning. Spotlighter evaluates each visual token’s activation from both sample-wise and semantic-wise perspectives and retains only the top-scoring tokens for downstream prediction. A class-specific semantic memory bank of learned prototypes refines this selection, ensuring semantic representativeness and compensating for discarded features. To further prioritize informative signals, we introduce a two-level ranking mechanism that dynamically weights token–prototype interactions. Across 11 few-shot benchmarks, Spotlighter outperforms CLIP by up to 11.19\%  in harmonic mean accuracy and achieves up to 0.8K additional FPS, with only 21 extra parameters. These results establish Spotlighter as an effective and scalable baseline for prompt tuning.
Code for our method will be available at \href{https://github.com/greatest-gourmet/Spotlighter}{{https://github.com/greatest -gourmet/Spotlighter}}.
\end{abstract}

\section{Introduction}
Recent advances in vision-language models have demonstrated remarkable capabilities in prompt tuning, particularly through approaches like CLIP~\cite{radford2021learning} that achieve robust cross-modal semantic alignment. These methods have demonstrated impressive performance in tasks such as open-domain recognition~\cite{cheng2024yolo}, fine-grained categorization~\cite{zhu2022dual}, and long-tailed distribution scenarios~\cite{liu2022long}, leading to breakthroughs in practical applications, including intelligent surveillance and medical image analysis. The superior performance of vision-language models primarily originates from their ability to learn discriminative joint embeddings that enable precise cross-modal alignment, a fundamental driver of continual model enhancement.

The alignment between visual and textual feature spaces enables effective classification, with ongoing research continuously enhancing representation quality through techniques like prompt learning~\cite{zhu2023prompt,xu2025progressive} and feature enhancement~\cite{sun2023eva,choi2025goal}. However, existing methods face two primary challenges: (1) Feature noise interference: Redundant or weakly relevant components within the aligned features introduce noise, undermining the contribution of semantically critical information~\cite{zhu2025weakclip}. (2) Computational efficiency bottleneck: Full-scale feature interactions across the entire representation space result in unnecessary computational burdens and higher costs in practical applications~\cite{khattakMaPLe}.
\begin{figure}
 \centering
  \includegraphics[width=1.0 \linewidth]{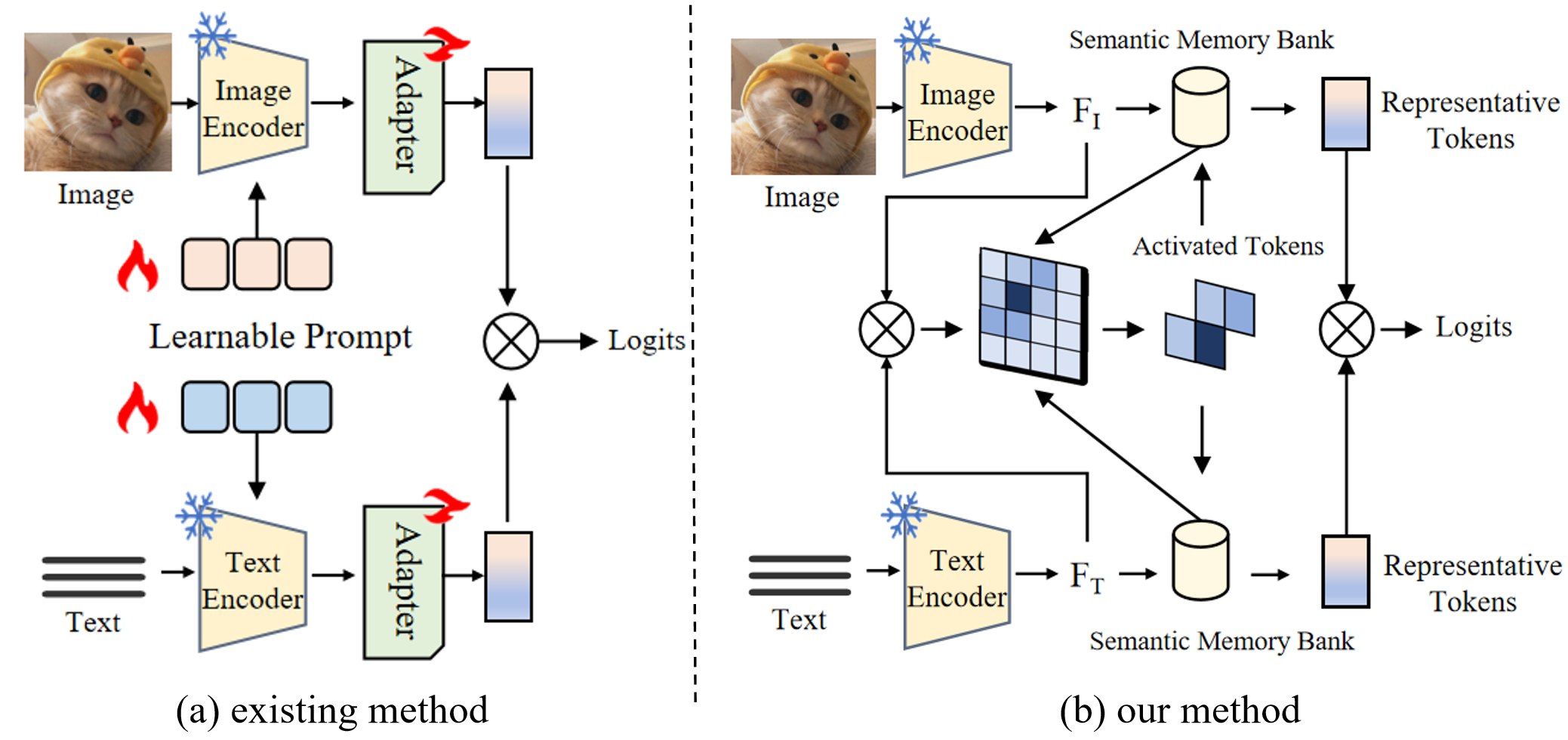}
  \vspace{-1.7em}
  \caption{Comparison with other methods. (a) Learnable prompts or adapters are applied to learn multimodal complex semantic information. (b) Activated and Representative token selection improves inference efficiency by mitigating noise and redundant features. }
  \label{fig:COM}  
\vspace{-1em}
\end{figure}

To address these challenges, prior works~\cite{huang2023structure,yang2025clip} have proved that during CLIP’s encoding process for effective image-text alignment, the model inherently captures a mixture of semantic signals. Since the image and text encoders operate independently, they are designed to cover a wide range of possible semantics.
This results in varied importance across different parts of the feature representation concerning specific classification goals.
Crucially, only a subset of these features contributes meaningfully to cross-modal alignment, while the rest may introduce redundancy.
Therefore, performing the sample-level evaluation of feature importance enables us to selectively emphasize critical features and suppress irrelevant ones, enhancing accuracy and efficiency. Existing approaches~\cite{khattakMaPLe,li2024vision,Khattak_2023_ICCV} employ global learnable text-image prompts or lightweight adapters in frozen layers to capture semantic information, as shown in Fig.\ref{fig:COM}(a), yet have not thoroughly explored the synergistic optimization between features representations and computational efficiency, leaving this as an open area for further research.
 
Based on the above analysis, we revisit cross-modal feature alignment in few-shot image classification and propose a simple yet effective model, Spotlighter, which achieves a favorable balance between accuracy and computational efficiency.
The key idea is to identify and retain sparse but highly representative feature tokens while discarding redundant ones.
Specifically, 
we evaluate each token's cross-modal semantic relevance from both sample-wise and semantic-wise perspectives, quantified as an activation score. Only a few highly activated tokens are retained for prediction, while the rest are discarded as redundant.
Unlike standard methods\cite{zou2023diffcr} that enhance typical tokens while keeping all, we strengthen representative tokens while discarding non-informative ones for greater efficiency.
To guide this selection, we introduce a semantic memory bank that stores a set of class-specific semantic prototypes. These prototypes help refine class boundaries during token activation, ensuring that the selected features are both semantically representative and capable of compensating for potentially missing information from discarded regions.
Furthermore, recognizing the varying contributions of activated features to classification, we introduce a two-level ranking mechanism over the prototypes. This mechanism dynamically adapts to the activation distribution of each sample, allowing the model to prioritize more informative features. The final representative features for prediction are then formed by fusing features with their corresponding prototypes according to their activation levels.

Extensive experiments conducted across 11 benchmark datasets demonstrate the effectiveness of our proposed method. Compared to CLIP~\cite{radford2021learning} and CLIPFit~\cite{li2024vision}, our approach achieves consistent improvements in both harmonic mean accuracy (HM) and computational speed, with an improvement of 11.19\% / 3.86\% in HM score and 0.8K/3.8K more FPS, respectively.
Remarkably, these gains come at the cost of only \textbf{21} additional parameters, highlighting the efficiency and scalability of our design.

Our main contributions are lies in:
\begin{itemize}
\vspace{-0.5em}
 \item We investigate the role of representative feature mining in prompt tuning, highlighting its dual benefits in improving both prediction accuracy and computational efficiency.
\vspace{-0.5em}
 \item 
 We propose Spotlighter, which selects the most activated tokens and enhance them via a semantic memory bank to form a compact yet informative representative feature set.
 \vspace{-0.5em}
  \item With only 21 additional parameters, our method boosts accuracy by 11.19\% and inference speed by 0.8K FPS over CLIP, establishing a strong, scalable baseline for prompt tuning.
\end{itemize}

\section{Related Works} 
\subsection{Pre-trained Vision-Language Models}
Large Language Models (LLMs) like GPT-3~\cite{brown2020language}, GPT-4~\cite{achiam2023gpt}, LLaMA~\cite{touvron2023llama}, and Deepseek~\cite{lu2024deepseek} exhibit robust zero-shot transfer capabilities in NLP tasks.
Nowadays,modern vision-language models (VLMs), enhanced by natural language supervision, excel in zero-shot/few-shot learning through large-scale image-text pretraining, as seen in contrastive learning-based models like ALIGN~\cite{li2021align} and CLIP~\cite{radford2021learning}.
Leveraging their formidable language-aligned visual representations and strong generalization, these models excel in diverse downstream tasks, such as object detection~\cite{zhang2022glipv2,gu2021open} and semantic segmentation~\cite{zhou2022zegclip,licascade}.
However, VLMs face significant challenges in degrading critical semantic information due to the redundant or weakly relevant components within the aligned features~\cite{zhu2023prompt,khattak2023self}.
Spotlighter enhances semantics and boosts efficiency through the hierarchical removal of useless components.
\subsection{Prompt Tuning}
Prompt learning adapts pre-trained models to downstream few-shot tasks via prompt-based reformulation, mitigating domain gaps and leveraging prior knowledge.
Early approaches~\cite{zhou2022cocoop,zhou2022coop,yao2023visual} relied on manually crafting templates based on prior human knowledge.
Later, MaPLe~\cite{khattakMaPLe}, PromptSRC~\cite{Khattak_2023_ICCV} concentrate on aligning visual-textual prompts jointly while adapter-based approaches~\cite{zhang2022glipv2,farina2025rethinking,lu2025variational,kim2024efficient,li2024vision} extend via context-aware prompt tuning using lightweight adapters in transformer layers.
Despite their success, these models often optimize prompts at coarse granularity, missing subtle visual cues and limiting cross-category generalization.
To solve this problem, ArGue~\cite{tian2024argue}, LLaMP~\cite{chiang2024llamp}, Texttrefiner~\cite{xie2024textrefiner} and SAR~\cite{jung2025learning} fill semantic gaps caused by noise through external LLMs or internal knowledge injection.
However, these methods require substantial memory consumption.
We enhance semantic information through multi-level feature tokenization while reducing large-scale feature interactions.
 
\begin{figure*}
 \centering
  \includegraphics[width=0.94\linewidth]{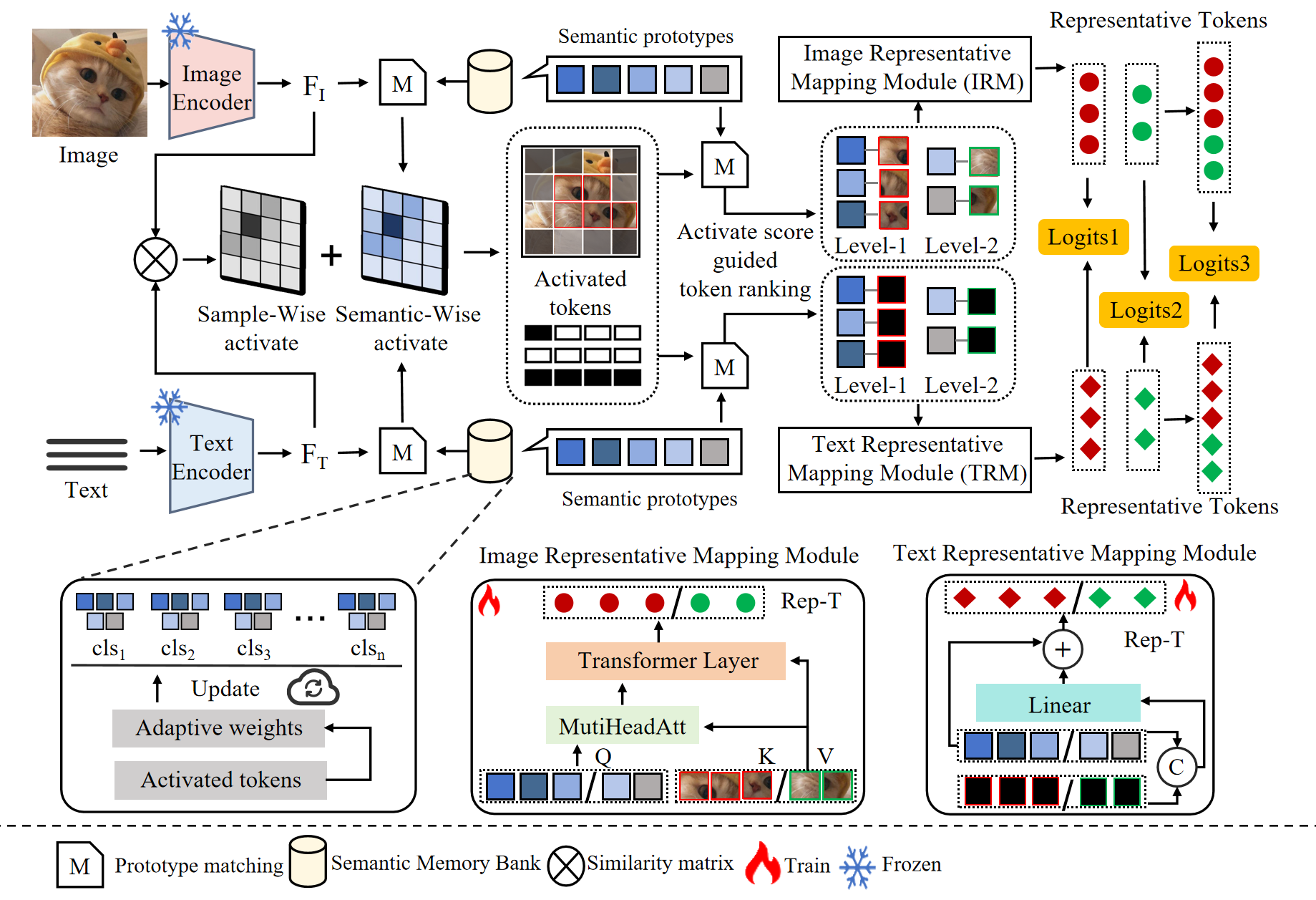}
  \caption{Overview of SpotLighter. The visual and textual features first compute sample-wise activations via a similarity matrix, which are fused with semantic-wise activations from prototype matching in the semantic memory bank. The combined activations yield $k$ tokens with the highest scores that are further refined through score-based stratification and processed by TIRM to obtain representative tokens. }
  \label{fig:TCP}
\end{figure*}

\section{Method}  
\subsection{Overview}
Vision-Language Models (VLMs), such as CLIP, leverage aligned image-text representations learned in a shared embedding space, offering advantages in few-shot image classification tasks.
Building on prior work, we adopt CLIP as our foundational model, with a key overview below.
CLIP consists of an image encoder, labeled as ${E}_I$, and a text encoder referred to as ${E}_T$.
Let $D = \{ (\boldsymbol{x}_i, \boldsymbol{t}_i)\}_{i = 1}^b$ represents the sampled batch, where $\boldsymbol{x}_i$ denotes the image input, $\boldsymbol{t}_i$ denotes the associated caption and b is the batch size.
Both encoders employ a feature extraction backbone followed by a projection layer that maps multi-modal inputs to a unified embedding space.
The image encoder encodes image $\boldsymbol{x}_i$ into $\boldsymbol{F}_I$, and text $\boldsymbol{t}_i$ into $\boldsymbol{F}_T$, \emph{i.e.},
\begin{equation}
  \boldsymbol{F}_I = E_{\mathrm{I}} (\boldsymbol{x}_i),\quad \boldsymbol{F}_T = E_{\mathrm{T}} (\boldsymbol{t}_i).
\end{equation}
During the training phase, a contrastive loss is employed to maximize the cosine similarity between them for alignment.
When testing, after getting the image feature $\boldsymbol{F}_I$ for image $\boldsymbol{x}_i$, the class $\boldsymbol{c}$ it belongs to is calculated by:
\begin{equation}
p (c) = \frac{\exp (\cos (\boldsymbol{T_c}, \boldsymbol{F}_{I}) / \tau)}{\sum_{j = 1}^{K} \exp (\cos (\boldsymbol{T}_j, \boldsymbol{F}_{I}) / \tau)},
\label{eq:prob_formula}
\end{equation}
where $\tau$ is a temperature parameter for scaling the softmax function, $\boldsymbol{T}_j$ is text embedding of class $\boldsymbol{j}$ and $\cos (\cdot, \cdot)$ denotes the cosine similarity function.
It is worth noting that CLIP aligns images and text by encoding them separately, but many features are noisy or redundant, thus extracting only the most relevant cross-modal features is necessary.
\subsection{Spotlighter}
To address the aforementioned challenges, we propose a plug-and-play method that selects a compact set of highly representative tokens. This strategy aims to suppress noise from redundant features and mitigate the computational overhead in the representative mining process.
Our method, Spotlighter, identifies activated tokens by leveraging a well-established paradigm from classical computer vision: intermediate-layer activations in visual networks naturally encode semantically salient and fine-grained visual concepts localized in specific image regions~\cite{zeiler2014visualizing, selvaraju2017grad, kim2022vit}. To enhance this capability, we introduce a Semantic Memory Bank, which facilitates the selection of representative tokens enriched with deeper semantic information.
By integrating feature activation with representative token extraction, Spotlighter captures rich semantics in a compact representation. An overview of the proposed framework is illustrated in Figure~\ref{fig:TCP}, and will be discussed below in detail.

\noindent\textbf{Feature Activation. }
To distill the most representative features across visual and textual modalities, we first evaluate each token's activation level in cross-modal semantic alignment for a given sample. These activation scores reflect the information distribution critical for prediction.
To obtain reliable activation scores, we compute them at both the sample and semantic levels. The sample-level score reveals cross-modal alignment between image-text pairs~\cite{selvaraju2017grad,wang2020score}, derived by computing the similarity between visual features $\boldsymbol{F}_I$ and textual features $\boldsymbol{F}_T$.
For semantic-level activation scores, we focus on capturing fine-grained semantic boundaries to enhance the representativeness of activated features.
To achieve this, we construct a set of prototypes~\cite{snell2017prototypical} for each semantic category, stored in a Semantic Memory Bank (SMB) ${U}\in\mathbb{R}^{k \times c}$, where $k$ is the number of prototypes and $c$ denotes the number of classes.
For prototype initialization, we employ class-name text embeddings as seed vectors. Ablation studies demonstrate that this text-guided approach outperforms random initialization.  
During training, we match each image feature $\boldsymbol{F}_I$ against all semantic prototypes $U$ in SMB to identify the most relevant semantic category $U_c$ ,
\begin{equation}
  {U}_c= \mathrm{argmax}\frac{\exp (\cos ( \boldsymbol{F}_{I},  {U}_j))}{\sum_{j = 1}^{C} \exp (\cos (\boldsymbol{F}_{I},  {U}_j))}.
  \label{equ1}
\end{equation}
We then compute semantic-level activation scores by comparing each prototype against both visual and textual similarity.
The final activation score is obtained by aggregating both sample-level and semantic-level activation scores.
Experimental evidence confirms that highly activated tokens offer more discriminative signals for sample prediction. We preserve only the most $k$ activated tokens $tok^{act}$ as classification evidence while treating the remaining features as redundant noise.
Notably, we continuously update the Semantic Memory Bank throughout the training process.
Firstly, we assign each of $tok^{act}$ to the corresponding prototype stored in $U$ by a softmax function to get the probability:
\begin{equation}
  \boldsymbol{D}_{i,j} = \frac{\exp (\cos ( {tok}_i^{act},  {U}_j))}{\sum_{j = 1}^{K} \exp (\cos ( {tok}_i^{act},  {U}_j))}.
  \label{equ1}
\end{equation}
Then, we assign by the highest probability as:
\begin{equation}
 \boldsymbol{U_j} = \left\{ i \, \middle\vert \, \underset{k}{\mathrm{argmax}} \boldsymbol{D}_{i,k} = j \right\}.
\end{equation}
Later, we will update the prototype in the Bank:
\begin{equation}
\label{beta}
  \boldsymbol{U}_j = \beta \cdot \boldsymbol{U}_j + (1 - \beta) \sum_{i \in U_j} \boldsymbol{D}_{i,j} \cdot  {tok}_i^{act},
\end{equation}
with $\beta$ representing the momentum coefficient.
To further ensure the effectiveness of the activated tokens, we calculate the similarity between the final $U$ of each class and sample-wise activation tokens:
\begin{equation}
\label{eq:local}
L_{local} = \text{CE}\left (\cos_{local} (U, tok^{act}), y\right).
\end{equation}
This local loss minimizes feature selection subjectivity through activation values, enhancing cross-modal knowledge transfer to compensate for limited pre-training interaction.

\noindent\textbf{Extraction of Representative Tokens. }
To compensate for potential semantic loss from discarded inactive regions, we fuse activated features with their corresponding semantic prototypes to obtain representative features for image classification. 
Given the varying predictive contributions of the activated features, we aim to guide the model to focus more on semantically relevant components by performing dynamic feature matching between the semantic prototype and the activated features of the current sample, and then stratifying them into two tiers based on activation scores. 
We then stratify them into two tiers based on their activation scores, denoted as $tok^{lev1}$ and $tok^{lev2}$.
To guide the model toward category-essential features, we progressively feed the prototypes and their matched activated tokens $tok^{lev1}$ and $tok^{lev2}$ respectively into the Image Representative Mapping Module (IRM) and the Text Representative Mapping Module (TRM), generating two sets of discriminative cross-modal representations.
In IRM, the activated tokens $tok^{lev_{i}}$ $(i=1,2)$ serve as both the key ${K}$ and the value ${V}$, while the corresponding prototypes $U$ serve as the query ${Q}$:
\begin{equation}
\small
  \boldsymbol{T} = \text{MultiHead} (\text{LN} (\boldsymbol{U}), \text{LN} ({ {tok}^{lev_{i}}}),\text{LN} ({ {tok}^{lev_{i}}})) + \boldsymbol{U}, 
\end{equation}
\begin{equation}
\small
    \widehat{\boldsymbol{U}} = \text{FFN} (\text{LN} (\boldsymbol{T})) + \boldsymbol{T},
    \label{loss2}
\end{equation}
where \text{MultiHead ($\cdot$)} and \text{FFN ($\cdot$)} follow the standard transformer, respectively representing multi-head attention and feed-forward neural network.
Subsequently, the fused token  $\widehat{\boldsymbol{U}}$ is concatenated with the ${tok}^{act}$ and processed through a transformer layer to obtain representative visual tokens
 \begin{equation}
   [ {tok}_v^{rep}, { {tok}}^{lev_i}] = \theta_i ([\widehat{\boldsymbol{U} }, { {tok}}^{lev_i}],
 \end{equation}
 where $[\cdot, \cdot]$ refers to the concatenation of each token and $\theta$ is the pre-trained transformer layer.
Meanwhile, for TRM, we begin by matching each original text token ${ {tok}_t^{ori}}$ with corresponding activated tokens using Eq.\ref{equ1} to get probability $W_{i,j}$.
Following this, we generate the final representative text tokens $tok_t^{rep}$ for activated token $i$ and utilize a residual-connected linear layer to fuse dual feature streams, where $\alpha$ is the coefficient hyperparameter:
\begin{equation}
\label{alpha}
\small
  { {tok}}_{i,t}^{rep} = \alpha \cdot \text{Linear} ([ {tok}_{t,i}^{ori} , \sum_{j = 1}^{k} W_{i,j} \cdot  {tok}_j^{lev_{i}}]) +  {tok}_{t,i}^{ori}.
\end{equation}
Notably, TRM processes text features through a single linear layer -- a deliberately simplified design contrasting with IRM's sophisticated image feature processing. Complete implementation details appear in Appendix~\ref{map}.
Then we concatenate the tokens achieved by Level-1 and Level-2 as ${tok}_{v}^{rep}$ and ${tok}_{t}^{rep}$.
Moreover, we posit that the set of high-activation-score features contains more discriminative information for classification.
Thus, we formulate $ \mathcal{L}_{cls}^{low}$ and $ \mathcal{L}_{cls}^{high}$to ensure independent classification capability for both feature sets, while reconstructing the  $ \mathcal{L}_{cls}^{gra}=\mathcal{L}_{cls}^{low}+\mathcal{L}_{cls}^{high}$ to prioritize high-representative features.
The way to calculate loss is similar to Eq.\ref{loss1}.

\subsection{Training and Inference}
Throughout the training process, we maintain the conventional CLIP architecture while employing contrastive loss as our fundamental classification objective, mathematically expressed as:
\begin{equation}
\small
  \mathcal{L}_{cls} = -\log\frac{\exp (\cos ( {tok}_{v}^{rep} ,  {tok}_{t}^{rep})/\tau)}{\sum_{j = 1}^{C} \exp (\cos ( {tok}_{v}^{rep},  {tok}_{t,i}^{rep})/\tau)}.
  \label{loss1}
\end{equation}
 
Beyond the standard contrastive loss formulation, we augment our module with a textual regularization loss and a visual KL loss, respectively:
\begin{equation}
  \mathcal{L}_{reg}^{text} = | {tok}_{t}^{ori} -  {tok}_{t}^{rep}|,
\end{equation}
\begin{equation}
  L_{KL}^{visual} = \text{KL} ( {tok}_{v}^{rep}, {tok}_{v}^{ori} ),
\end{equation}
where $KL (\cdot, \cdot)$ represents Kullback-Leibler divergence and $ {tok}_{t,v}^{ori}$ is the original text and visual tokens achieved by pre-trained models.
The $\mathcal{L}_{reg}^{text}$ can mitigate overfitting in VLMs fine-tuning with limited training data, while $ L_{KL}^{visual}$ ensures useful image tokens exhibiting strong alignment with the original pre-trained feature space.
Then the total loss can be calculated:
\begin{equation}
\small
\label{total}
  \mathcal{L} = \mathcal{L}_{cls}  + \lambda_1*\mathcal{L}_{cls}^{gra}+\lambda_2*\mathcal{L}_{reg}^{text}+\lambda_3 * (\mathcal{L}_{KL}^{visual}+\mathcal{L}_{local}),
\end{equation}
where $\lambda_1$, $\lambda_2$, $\lambda_3$ are hyper-parameters used to balance the various loss terms.
In all, we only need to train \textbf{the parameters in Eq. \ref{loss2} and Eq. \ref{alpha}}, thus improving training efficiency.

During inference, we compute the final prediction scores using the fused cross-modal representations from both visual and textual features:
\begin{equation}
\small
  y = \underset{i}{\text{arg}\,\text{max}} \frac{\exp (\cos ({tok}_{t}^{rep}, {tok}_{r}^{rep})/\tau)}{\sum_{j = 1}^{C} \exp (\cos ({tok}_{t}^{rep}, {tok}_{r}^{rep})/\tau)}.
\end{equation}
Unlike existing approaches that rely on redundant remaining tokens after alignment, our method simply performs inference by the most representative tokens, thus mitigating noise-induced semantic degradation while reducing high-dimensional feature interactions in the representation space. 

\begin{table*}[t] 
  
  \caption{Comparison with other methods on base-to-new generalization with 16-shot.} 
  \label{tab:base-to-new-generalization}
  \centering
  \resizebox{\textwidth}{!}{ 
  \begin{tabular}{l|ccc|ccc|ccc|ccccccc} 
    \toprule
    \multirow{2}{*}{Method} & \multicolumn{3}{c}{Average} & \multicolumn{3}{c}{ImageNet} & \multicolumn{3}{c}{Caltech101} & \multicolumn{3}{c}{OxfordPets} \\
    \cmidrule (lr){2 - 4} \cmidrule (lr){5 - 7} \cmidrule (lr){8 - 10} \cmidrule (lr){11 - 13}
    & Base & Novel & HM & Base & Novel & HM & Base & Novel & HM & Base & Novel & HM \\
    \midrule
    CLIP & 69.34 & 74.22 & 71.70 & 72.43 & 68.14 & 70.22 & 96.84 & 94.00 & 95.40 & 91.17 & 97.26 & 94.12 \\
    CoOp & 82.69 & 63.22 & 71.66 & 76.47 & 68.78 & 71.92 & 98.00 & 89.81 & 93.73 & 93.67 & 95.29 & 94.47 \\
  
    PromptSRC & 84.26 & 76.10 & 79.97 & 77.60 & 70.73 & 74.01 & 98.10 & 94.03 & 96.02 & 95.33 & 97.30 & 96.30 \\
    MaPLe & 82.28 & 75.14 & 78.55 & 76.66 & 70.54 & 73.47 & 97.74 & 94.36 & 96.02 & 95.43 & 97.76 & 96.58 \\
    CLIPFit &83.72 &74.84 &79.03 &76.20 &70.17 &73.06 &98.30 &93.70 &95.94 &95.23 &97.13 &96.17\\
    PromptKD & 84.11 & 78.28 & 81.09 & \textbf{77.63} & 70.96 & 74.15 & 98.31 & 96.29 & 97.29 & 93.42 & 97.44 & 95.39 \\
    CoOp $w$/ TextRefiner & 79.74 & 74.32 & 76.94 & 76.84 & 70.54 & 73.56 & 98.13 & 94.43 & 96.24 & 95.27 & 97.65 & 96.45 \\
    PromptKD $w$/ TextRefiner & 85.22 & 79.64 & 82.33 & 77.51 & 71.43 & 74.38 & 98.52 & 96.52 & 97.51 & 95.60 &\textbf{ 97.90} & 96.74 \\
    \midrule
    \rowcolor{blue!4}
    CoOp $w$/ \textit{Spotlighter} &81.74 &75.80 &78.66 &76.74 &70.68 &73.58 &98.13 &94.51 &96.29 &\textbf{97.40} &97.73 &
    \textbf{97.56}\\
    \rowcolor{blue!4}
    PromptKD $w$/ \textit{Spotlighter} &\textbf{85.65} &\textbf{80.46} &\textbf{82.89} &77.62&\textbf{71.71} &\textbf{74.55} &\textbf{98.86} &\textbf{96.74} &\textbf{97.79} &96.48 &97.75 &97.11 \\
    \midrule
    \multirow{2}{*}{Method} & \multicolumn{3}{c}{StanfordCars} & \multicolumn{3}{c}{Flowers102} & \multicolumn{3}{c}{Food101} & \multicolumn{3}{c}{FGVCAircraft} \\
    \cmidrule (lr){2 - 4} \cmidrule (lr){5 - 7} \cmidrule (lr){8 - 10} \cmidrule (lr){11 - 13}
    & Base & Novel & HM & Base & Novel & HM & Base & Novel & HM & Base & Novel & HM \\
    \midrule
    CLIP & 63.37 & 74.89 & 68.65 & 72.08 & 77.80 & 74.83 & 90.10 & 91.22 & 90.66 & 27.19 & 36.29 & 31.09 \\
    CoOp & 78.12 & 60.40 & 68.13 & 97.60 & 59.67 & 74.06 & 88.33 & 82.26 & 85.19 & 40.44 & 22.30 & 28.75 \\
    PromptSRC & 78.27 & 74.97 & 76.58 & 98.07 & 76.50 & 85.95 & 90.67 & 91.53 & 91.10 & 42.73 & 37.87 & 40.15 \\
    MaPLe & 72.94 & 74.00 & 73.47 & 95.92 & 72.46 & 82.56 & 90.71 & 92.05 & 91.38 & 37.44 & 35.61 & 36.50 \\
    CLIPFit &78.80 &73.87 &76.26 &96.83 &73.53 &83.59 &90.60 &91.33 &90.96 &42.47 &33.47 &37.43\\
    PromptKD & 80.48 & 81.78 & 81.12 & 98.69 & 81.91 & 89.52 & 89.43 & 91.27 & 90.34 & 43.61 & 39.68 & 41.55 \\
    CoOp $w$/ TextRefiner & 71.40 & 70.90 & 71.15 & 95.92 & 74.33 & 83.76 & 90.88 & 91.43 & 91.15 & 35.35 & 35.87 & 35.61 \\
    PromptKD $w$/ TextRefiner & 80.91 & 81.83 & 81.37 & 99.30 & 82.91 & 90.37 & 91.42 & 92.71 & 92.06 & 45.01 & 40.12 & 42.42 \\
    \midrule
    \rowcolor{blue!4}
     CoOp $w$/\textit{Spotlighter} &70.09 &69.97 &70.03 &95.10 &74.47 &83.53 &\textbf{93.63} &91.51 &\textbf{92.56} &39.00 &36.54 &37.72 \\
     \rowcolor{blue!4}
       PromptKD $w$/\textit{Spotlighter} &\textbf{81.62} &\textbf{82.15} &\textbf{81.88} &\textbf{99.36} &\textbf{83.47} &\textbf{90.72} &91.86 &\textbf{92.93} &92.39 &\textbf{46.35} &\textbf{40.68} &\textbf{43.33}\\
     \midrule
    \multirow{2}{*}{Method} & \multicolumn{3}{c}{SUN397} & \multicolumn{3}{c}{DTD} & \multicolumn{3}{c}{EuroSAT} & \multicolumn{3}{c}{UCF101} \\
    \cmidrule (lr){2 - 4} \cmidrule (lr){5 - 7} \cmidrule (lr){8 - 10} \cmidrule (lr){11 - 13}
    & Base & Novel & HM & Base & Novel & HM & Base & Novel & HM & Base & Novel & HM \\
    \midrule
    CLIP & 69.36 & 75.35 & 72.23 & 53.24 & 59.90 & 56.37 & 56.48 & 64.05 & 60.03 & 70.53 & 77.50 & 73.85 \\
    CoOp & 80.60 & 65.89 & 72.51 & 79.44 & 41.18 & 54.24 & 92.19 & 54.74 & 68.69 & 84.69 & 56.05 & 67.46 \\
    PromptSRC & 82.67 & 78.47 & 80.52 & 83.37 & 62.97 & 71.75 & 92.90 & 73.90 & 82.32 & 87.10 & 78.80 & 82.74 \\
    MaPLe & 80.82 & 78.70 & 79.75 & 80.36 & 59.18 & 68.16 & \textbf{94.07} & 73.23 & 82.35 & 83.00 & 78.66 & 80.77 \\
    CLIPFit &81.97&78.17 &80.02 &81.97 &63.50 &71.56 &93.33 &71.07 &80.69 &85.23 &77.30 &81.07\\
    PromptKD & 82.53 & 80.88 & 81.70 & 82.86 & 69.15 & 75.39 & 92.04 & 71.59 & 80.54 & 86.23 & 80.11 & 83.06 \\
    CoOp $w$/ TextRefiner & 80.96 & 76.49 & 78.66 & 75.35 & 58.09 & 65.60 & 74.57 & 72.82 & 73.68 & 82.52 & 75.01 & 78.59 \\
    PromptKD $w$/ TextRefiner & 83.02 & 80.50 & 81.74 & 83.91 & 71.01 & 76.92 & 92.99 & 79.22 & 85.55 & 89.20 & 81.90 & 85.39 \\
    \midrule
    \rowcolor{blue!4}
    CoOp $w$/ \textit{Spotlighter} &81.78 &75.17 &78.48 &76.04 &58.69 &66.18 &85.01 &82.13 &83.85 &86.27 &\textbf{82.47} &84.34 \\
    \rowcolor{blue!4}
    PromptKD $w$/ \textit{Spotlighter} &\textbf{83.15} &\textbf{81.06} &\textbf{82.09} &\textbf{83.94} &\textbf{71.92} &\textbf{77.47} &93.17 &\textbf{84.51} &\textbf{88.63} &\textbf{89.72} &82.16 &\textbf{85.77}\\
    \bottomrule
  \end{tabular}
  }
\end{table*}

\section{Experiments}
\subsection{Experimental Settings}
\noindent\textbf{Datasets.}
We employ the conventional approach used in previous studies~\cite{zhou2022cocoop,khattakMaPLe} to conduct the base-to-new and few-shot on 11 benchmarks, \emph{i.e.,} ImageNet~\cite{DengDSLL009}, Caltech~\cite{Fei-FeiFP07}, OxfordPets~\cite{ParkhiVZJ12}, StanfordCars~\cite{Krause0DF13}, Flowers~\cite{NilsbackZ08}, Food101~\cite{BossardGG14}, FGVCAircraft~\cite{MajiRKBV13}, EuroSAT~\cite{HelberBDB19}, UCF101~\cite{abs-1212-0402}, DTD~\cite{CimpoiMKMV14}, and SUN397~\cite{XiaoHEOT10}. For cross-dataset generalization, we experiment on ImageNet-V2~\cite{recht2019imagenet}, ImageNet-Sketch~\cite{wang2019learning}, ImageNet-A~\cite{hendrycks2021natural} and ImageNet-R~\cite{hendrycks2021many}. 
The Implementation Details will be discussed in Apeendix~\ref{detal}.
\begin{table}[t]
\caption{Comparison with other methods on the few-shot learning setting with average accuracy.
We plug our method in PrompKD.}
\label{few-shots}
\centering
\small
\begin{tabular}{c|ccccc}
\hline
\multirow{2}{*}{Method} & \multicolumn{5}{c}{Shot} \\
& 1 & 2 & 4 & 8 & 16 \\
\hline
CLIP  &45.12 &54.63 &65.24 &66.87 &71.70   \\
CoOP & 68.09 & 70.13 & 73.59 & 76.45 & 79.01 \\
PromptSRC &72.32 &75.28 & 78.35 &80.69 &82.87\\
MaPLe &61.79 &65.28 &70.66 &73.82 &78.55 \\
CLIPFit &72.32 &74.39 &77.18 &79.03 &81.27 \\
PromptKD &72.47&75.19&78.46 &79.56 &81.09\\
\midrule
\rowcolor{blue!10}
 $w$/ \textit{Spotlighter} & \textbf{72.53} & \textbf{75.76} & \textbf{78.80} & \textbf{81.81} & \textbf{85.65} \\
\hline
\end{tabular}
\end{table}
\begin{table}[h]
\setlength{\tabcolsep}{4pt} 
\caption{Comparison with other methods on cross-domain generalization with 16-shot.} 
\label{cross-domain}
  \centering
  \small 
  \begin{tabular}{l*{5}{c}}
    \toprule
    \multirow{2}{*}{Method} & \multicolumn{1}{c}{Source} & \multicolumn{4}{c}{Target} \\
    \cmidrule (lr){2-2} \cmidrule (lr){3-6} 
    \addlinespace[1pt] 
    & \multicolumn{1}{c}{ImageNet} & \multicolumn{1}{c}{-V2} & \multicolumn{1}{c}{-Sketch} & \multicolumn{1}{c}{-A} & \multicolumn{1}{c}{-R} \\
    \midrule
    CLIP & 66.73 & 60.83 & 46.15 & 47.77 & 73.96 \\
    CoOpOp & 71.02 & 64.07 & 48.75 & 50.63 & 76.18 \\
    PromptSRC & 71.27 & 64.35 & 49.55 & 50.90 & 77.80 \\
    \midrule
    CoOp & 71.51 & 64.20 & 47.99 & 49.71 & 75.21 \\
    \rowcolor{blue!10}
    $w$/ \textit{Spotlighter} & 72.12 & 66.17 & 49.32 & 49.81 & 76.59 \\
    MaPLe & 70.72 & 64.05 & 49.15 & 50.90 & 76.98 \\
    \rowcolor{blue!10}
    $w$/ \textit{Spotlighter} & \textbf{72.17} & \textbf{69.62} & \textbf{50.18} & \textbf{69.83} & \textbf{83.56} \\
    \bottomrule
  \end{tabular}
\end{table}
\begin{table}[h]
  \caption{Comparison of inference efficiency among existing methods on the ImageNet Dataset.}
\setlength{\tabcolsep}{4pt} 
  \label{efficiency}
  \centering
  \small
  \begin{tabular}{lcc|c}
    \toprule
    Method &Params & FPS & HM \\
    \hline
    CoOp &2048 & 9768.21 & 71.92 \\
    CoCoOp &35K  & 20.45 & 73.10 \\
    CLIPFit &44K & 8380.91 & 73.06\\
    LLaMP  &5.2M &1473.46 &74.48\\
    PromptKD & 2.5M & 12943.34 & 74.15 \\
    \midrule
    CoOp $w$/ \textit{Spotlighter} &\textbf{+21} &\textbf{+886.61} &\textbf{+1.66}\\
    PromptKD $w$/ \textit{Spotlighter} &\textbf{+21} &\textbf{+1813.52} &\textbf{+0.35}\\ 
    \bottomrule
  \end{tabular}
\end{table}
\noindent\textbf{Baselines.}
We compare with many state-of-the-art (SOTA) method, including CLIP~\cite{radford2021learning}, CoOp~\cite{zhou2022coop}, PromptSRC~\cite{zhu2023prompt}, MaPLe~\cite{khattakMaPLe}, CLIPFit~\cite{li2024vision}, PromptKD~\cite{li2024promptkd} and TextRefiner~\cite{xie2024textrefiner}.

\subsection{Comparison with State-of-art Methods}
\noindent\textbf{Base-to-Novel Generalization.}
Table~\ref{tab:base-to-new-generalization} presents the quantitative results of various methods in the base-to-novel generalization setting on 11 datasets.
Our method demonstrates significant capability in consistently enhancing the performance of existing approaches across all evaluation metrics (Base, New, and HM), outperforming competing methods.
Notably, Spotlighter significantly boosts CoOp's generalization capability on novel classes, achieving a remarkable accuracy improvement from 63.22\% to 75.80\%.
With PromptKD, Spotlighter achieves the best accuracy to 85.65\% on the base while improving the Novel to 80.46\%.
This verifies that after filtering out weakly relevant tokens, our model can reduce noise introduction while enhancing relevant semantic information, improving the model's generalization capability.
 
\noindent\textbf{Few-shot Classification. }
In the few-shot scenario, our method also performs well.
Following CLIP, we used 1/2/4/8/16-shot settings for training and calculated the accuracy on 11 datasets.
Table~\ref{few-shots} shows when compared with other methods, Spotlighter displays overall consistent improvement among all settings, demonstrating robustness and efficacy even in challenging low-data regimes.

\noindent\textbf{Cross-Datasets Generalization. }
Extending beyond standard benchmarks, we assess Spotlighter's cross-domain generalization on four established datasets. The results shown in Table~\ref{cross-domain} verify that in cross-data scenarios, Spotlighter can still show the best results after few-shot training on ImageNet, especially for ImageNet-A having \textbf{18.93\% } improvement.
This demonstrates through progressive refinement, even a limited set of representative tokens can retain sufficient semantic information.

\noindent\textbf{Efficiency.}
We further conduct a comparative analysis of inference efficiency, benchmarked on a single NVIDIA 4090 GPU using the officially released implementation.
As shown in Table~\ref{efficiency}, when plugging in Spotlighter, other methods achieve faster inference speeds.
Notably, with only \textbf{21 additional parameters}, Spotlighter not only attains the best performance in HM at the fastest inference speed. This efficiency gain is primarily due to using a compact set of semantically rich representative tokens, which substantially reduces the scale of feature interactions across the representation space, leading to a notable reduction in computational overhead.
\begin{table}[t]
\setlength{\tabcolsep}{1.8pt} 
\small
\centering
\caption{Ablation experiments on different optimization losses on ImageNet.}
\scalebox{1}{
\begin{tabular}{cccccc|ccc}
  \toprule
  $L_{cls}$ & $L_{local}$ & $L_{cls}^{low}$ & $L_{cls}^{high}$ &$L_{reg}^{text}$ &$L_{kl}^{visual}$&Base & Novel & HM \\ 
  \midrule
  \checkmark & & & & & &76.50 & 67.88 & 71.93 \\ 
  \checkmark & & & &\checkmark & &76.58 & 70.62 & 73.48 \\ 
  \checkmark & \checkmark &  & & & \checkmark &76.16 & 69.75 & 72.81 \\ 
  \checkmark & \checkmark &\checkmark & \checkmark & & & 76.24 & 69.88 & 72.92\\ 
  \checkmark & \checkmark & \checkmark & \checkmark &\checkmark & & 76.13 & 70.31 &73.10 \\ 
  \checkmark & \checkmark & \checkmark & \checkmark & & \checkmark&76.47 & 70.32 & 73.27 \\ 
  \midrule
    \rowcolor{blue!10}
    \checkmark & \checkmark & \checkmark  & & \checkmark & \checkmark& 76.98 &71.16 &73.96  \\ 
      \rowcolor{blue!10}
      \checkmark & \checkmark &   & \checkmark & \checkmark & \checkmark& 77.25 &  71.34&  74.18 \\ 
        \rowcolor{blue!10}
    \checkmark & \checkmark & \checkmark & \checkmark & \checkmark & \checkmark& \textbf{77.62} & \textbf{71.71} & \textbf{74.55} \\ 
  \bottomrule
\end{tabular}}
\label{tab:results5}
\end{table}
 \begin{table}[t]
\setlength{\tabcolsep}{4pt} 
    \centering
    \caption{Effects of  $tok^{lev1}$ and $tok^{lev2}$ in Inference.}
    \small
    \begin{tabular}{c|ccc|c}
        \toprule
        Method & Base & Novel & HM  &FPS\\
        \hline
        $tok^{lev1}$ & 75.28 & 70.16 & 72.63 & 221.57K\\
    $tok^{lev2}$ & 75.47 & 70.29 & 72.79 &216.32K \\
        $tok^{lev\_1+2}$ & \textbf{77.62} & \textbf{71.71} & \textbf{74.55}  &131.25K\\
        \bottomrule
    \end{tabular}
    \label{tab:resul}
\end{table}
\subsection{Ablation Experiments}
\noindent\textbf{Effects of Different Losses.}
In the training process, we introduced a variety of training losses, shown in Eq.\ref{total}.
Table~\ref{tab:results5} investigates the influence of these factors on the model's generalization capability. The introduced $L_{local}$ ensures the preservation of semantic information in useful tokens, while $L_{reg}^{text}$ and $L_{kl}^{visual}$ incorporate knowledge from original tokens and constrain fine-grained information utilization. Additionally,  $  \mathcal{L}_{cls}^{low}$ and $\mathcal{L}_{cls}^{high}$ enhance cross-modal interaction.
Empirical results show that combining multiple training objectives effectively balances adaptability and generalization, leading to improved overall performance.

\noindent\textbf{Effects of the Activated and Representative Tokens.}
\begin{figure}[t]
 \centering
\includegraphics[width=1.0\linewidth]{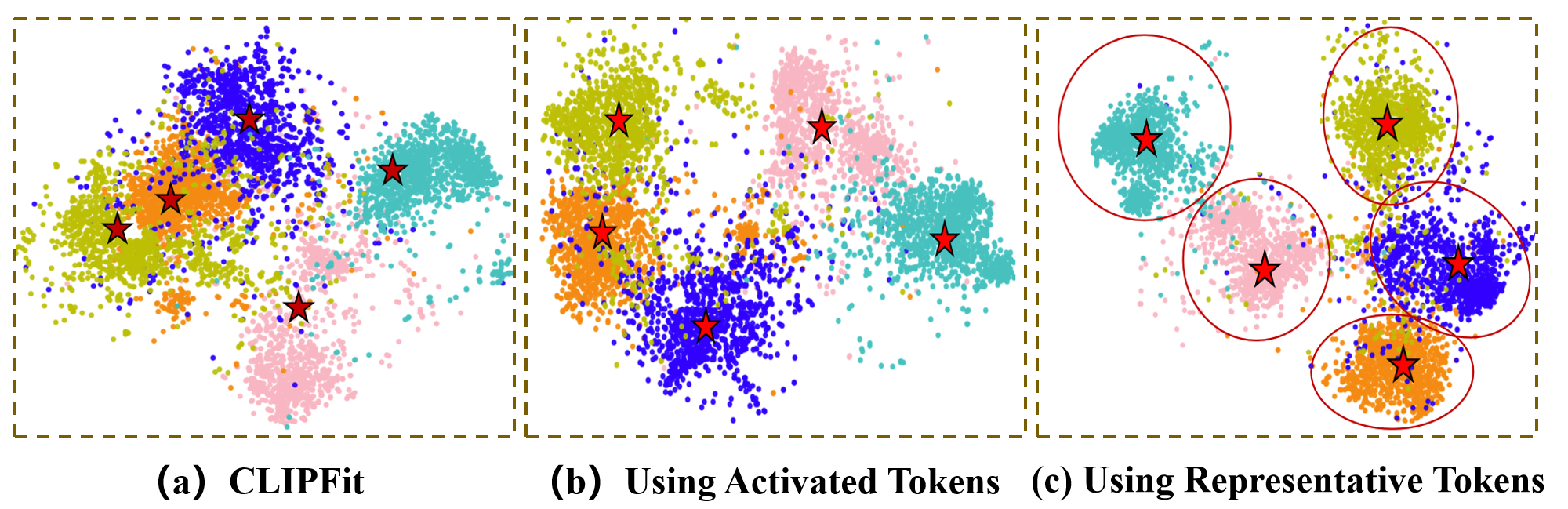}
  \caption{Visualization of the effect of the activated and representative tokens on the ImageNet dataset in few-shot learning via t-SNE.}
  \label{t-sne}
\end{figure}
To boost salient tokens' information density during aggressive pruning, we enhance the activated tokens to get representative tokens with ImageNet t-SNE~\cite{selvaraju2017grad} visualizations.
From Fig.\ref{t-sne}, we can observe that with Spotlighter, CLIPFit can have a much clearer separation of different class image features and more correct text features embedding in high-dimensional feature space, which contingents upon more granular stratification.
Additionally, representative tokens can have better distinguishing capability.

\noindent\textbf{Effects of the Two-Level Feature Activation.}
We stratify activated tokens by activation scores into Level-1 and Level-2 subsets, yielding more representative tokens for finer alignment and richer semantics.
From Table~\ref{tab:resul} and Table~\ref{tab:results5}, we observe that using only Level-1 or Level-2 tokens improves efficiency but sacrifices semantic coverage either in loss or inference.
Therefore, unifying levels for better cross-modal interaction is necessary.
\begin{table}[t]
  \centering
  \small
  \caption{Comparison of baseline methods with and without Spotlighter in an average of 16 datasets.}
  \label{plug}
  \begin{tabular}{lccc}
  
    \toprule
    Method & Base & Novel & HM \\
    \midrule
     
    PromptKD & 84.11 & 78.28 & 81.09 \\
    $w/$ \textit{Spotlighter} & \textbf{85.65}\textcolor{red}{\scriptsize{+1.54}} & \textbf{80.46}\textcolor{red}{\scriptsize{+2.18}} & \textbf{82.89}\textcolor{red}{\scriptsize{+1.80}} \\
    \midrule
    CLIPFit & 83.72 & 74.84 & 79.03 \\
    $w/$ \textit{Spotlighter} & \textbf{85.17}\textcolor{red}{\scriptsize{+1.45}} & \textbf{78.62}\textcolor{red}{\scriptsize{+3.78}} & \textbf{81.76}\textcolor{red}{\scriptsize{+2.73}} \\
    \midrule
    MaPLe & 82.28 & 75.14 & 78.55 \\
    $w/$ \textit{Spotlighter} & \textbf{83.29}\textcolor{red}{\scriptsize{+1.01}} & \textbf{77.45}\textcolor{red}{\scriptsize{+2.31}} & \textbf{80.26}\textcolor{red}{\scriptsize{+1.71}} \\
    \bottomrule
  \end{tabular}
\end{table}
\begin{table}[h]
\setlength{\tabcolsep}{4pt} 
    \centering
    \small
    \caption{Effects of the semantic action tokens.}
    \begin{tabular}{c|ccc}
        \toprule
        Method & Base & Novel & HM \\
        \hline
       \textit{ Spotlighter} $w/o$ Semantic Action& 77.52 & 71.43 &  74.35  \\
       \rowcolor{blue!10}
         \textit{ Spotlighter} $w$ Semantic Action &  77.62 &  71.71 &  74.55   \\
        \bottomrule
    \end{tabular}
    \label{tab:semantic}
\end{table}
\begin{table}[h]
\setlength{\tabcolsep}{4pt} 
    \centering
    \small
    \caption{Different ways for prototype initialization.}
    \begin{tabular}{c|ccc}
        \toprule
        Method & Base & Novel & HM \\
        \hline
      Random Initialization & 82.54 & 77.46 &  79.92  \\
      \rowcolor{blue!10}
      Text Embedding Seeds &  85.65 &  80.46 &  82.89   \\
        \bottomrule
    \end{tabular}
    \label{tab:initialization}
\end{table}
\noindent\textbf{Effects on Different Backbones.}    
To systematically examine the plug-and-play functionality of Spotlighter and demonstrate broad applicability, we implement the approach across multiple representative frameworks.
Shown in Table~\ref{plug}, all four methods exhibit significant improvement, confirming effectiveness and versatility. 
 


\noindent\textbf{Effects of the Semantic Activation.}
When computing activated tokens, we add the activation scores of the sample and the semantics.
We observe from Table~\ref{tab:semantic} that empowered by Semantic Activation Tokens, the sampled acquire richer and more discriminative semantic representations.
This is because the prototypes store the most salient information of each image category and are continuously refined through updates. Their integration with individual samples mitigates the effects of sample-level variance and information sparsity, ultimately leading to higher-quality activated tokens.

\noindent\textbf{Different Ways for Prototype Initialization.}
To capture fine-grained semantic boundaries and enhance the representational capacity of activated features, we construct prototypes for each semantic category. The table~\ref{tab:initialization} presents the average results of prototype initialization across 11 datasets, comparing Random Initialization with Text Embedding Seeds.
The results show that, with the guidance of Text Embedding Seeds, the prototypes better capture representative information of their corresponding categories, thereby facilitating a more effective construction of the semantic memory bank.
\section{Conclusion}
We introduce Spotlighter, a plug-and-play framework that revisits few-shot image classification from the perspective of representative token mining. By progressively selecting and categorizing informative tokens, Spotlighter effectively filters noise and reduces redundant feature interactions. Leveraging both activated tokens and representative tokens, the model enhances fine-grained cross-modal alignment with minimal parameter overhead. Extensive experiments across 11 benchmarks and diverse generalization settings show that Spotlighter consistently improves accuracy and efficiency over strong baselines. Our work highlights the importance of token-level selection and structured refinement for efficient and robust few-shot learning with vision-language models.

\section*{Acknowledgments}
This work is supported by the Major Projects of the National Natural Science Foundation of China(Grant No.72293580/72293583); supported by the Hainan Provincial Joint Project of Lian International Education Innovation Pilot Zone, Grant No: 624LALH002 ; supported by the Beijing Natural Science FouL221011.L221011.
 
\bibliography{anthology,custom}
\bibliographystyle{acl_natbib}
\clearpage
\appendix

\begin{center}
    { \textbf{Supplementary Material of \\ \textit{Spotlighter: Revisiting Prompt Tuning from a Representative Mining View  }}}
\end{center}

\section{Limitations }
Spotlighter is primarily designed for fine-tuning vision-language models in image classification and may not generalize well to other vision tasks such as object detection or image segmentation, where dense or spatially localized predictions are required. This limitation partly stems from the reduced number of final tokens, which may omit fine-grained spatial details essential for those tasks. Moreover, the effectiveness of our method relies on the presence of sufficiently discriminative representative tokens; performance degrades when such tokens are sparse or class boundaries are highly entangled, particularly in ultra-fine-grained settings. In future work, we plan to extend Spotlighter to dense prediction tasks by incorporating spatial-aware token selection and hierarchical refinement. We also aim to investigate adaptive token filtering strategies that dynamically adjust to data complexity and class granularity.

\section{Dataset Statistics}
To rigorously assess the effectiveness and generalization ability of our method, we performed extensive experiments on 11 standard benchmark datasets spanning multiple visual domains (Table~\ref{tab:dataset}). The selected datasets cover diverse recognition tasks including: ImageNet~\cite{DengDSLL009} for object classification; Caltech~\cite{Fei-FeiFP07} for natural object recognition; OxfordPets~\cite{ParkhiVZJ12} for fine-grained pet classification; StanfordCars~\cite{Krause0DF13} for vehicle categorization; Flowers~\cite{NilsbackZ08} for flower species identification; Food101~\cite{BossardGG14} for food classification; FGVCAircraft~\cite{MajiRKBV13} for aircraft recognition; EuroSAT~\cite{HelberBDB19} for satellite imagery analysis; UCF101~\cite{abs-1212-0402} for action recognition; DTD~\cite{CimpoiMKMV14} for texture classification; and SUN397~\cite{XiaoHEOT10} for scene understanding.In distribution shift experiments,we also introduce ImageNet-V2~\cite{recht2019imagenet}, ImageNet-Sketch~\cite{wang2019learning}, ImageNet-A~\cite{hendrycks2021natural} and ImageNet-R~\cite{hendrycks2021many}.These datasets are all to improve ImageNet test reliability. This comprehensive evaluation across multiple domains effectively demonstrates our approach's robustness and versatility in various scenarios.

\section{Implementation Details}
\label{detal}
We adopt ViT-B/16 CLIP model to conduct all of our experiments. We report both base and novel class accuracies along with their harmonic mean (HM)~\cite{xian2017zero}, with all metrics averaged across three independent runs. 
Here, base refers to the accuracy on seen classes, novel refers to the accuracy on unseen classes, and HM denotes the harmonic mean of the two. These metrics evaluate the model’s performance on seen, unseen, and their balance.
To ensure a fair comparison, final performance metrics are computed as the mean over three different random seeds. The experimental settings remain consistent with the original papers while the only modification is in the number of training epochs where CoOp is reduced to 15 epochs, while ClipFit and PromptKD are reduced to 30 epochs. The number of fusion coefficient $\alpha$ is 0.2 and momentum coefficient $\beta$ is 0.8 respectively.
For the hyper-parameters in the loss, we set $\lambda_1$, $\lambda_2$, $\lambda_3$ to
0.02, 20, 0.1 supported by empirical findings and fixed in different datasets to facilitate downstream tasks.
Furthermore, each category in the memory bank maintains only five lightweight prototypes of 512 dimensions, incurring an average overhead of merely 10KB per category, which is a reasonable trade-off for notable computational efficiency gains.
\begin{table*}[h]
\centering
\setlength{\tabcolsep}{3pt}
\caption{The detailed statistics of datasets used in our work.}
\label{tab:dataset}
\resizebox{\textwidth}{!}{
\begin{tabular}{l|c|c|c|c|c|c}
\toprule
\textbf{Datasets} & \textbf{Classes} & \textbf{Training Size} & \textbf{Validation Size} & \textbf{Testing Size} & \textbf{Tasks} & \textbf{Hand-crafted Prompt} \\
\midrule
ImageNet     & 1,000 & 1.28M & N/A  & 50,000 & General object recognition & "a photo of a [CLASS]."\\ 
Caltech      & 100  & 4,128 & 1,649 & 2,465 & General object recognition & "a photo of a [CLASS]." \\
EuroSAT      & 10  & 13,500 & 5,400 & 8,100 & Satellite image recognition &"a centered satellite photo of [CLASS]."  \\
SUN397       & 397  & 15,880 & 3,970 & 19,850 & Scene recognition &"a photo of a [CLASS]." \\
DTD       & 47  & 2,820 & 1,128 & 1,692 & Texture recognition & "[CLASS] texture." \\
UCF101     & 101  & 7,639 & 1,808 & 3,783 & Action recognition & "a photo of a person doing [CLASS]." \\
FGVCAircraft   & 100  & 3,334 & 3,333 & 3,333 & Fine-grained aircraft recognition & "a photo of a [CLASS], a type of aircraft." \\
OxfordPets    & 37  & 2,944 & 736  & 3,669 & Fine-grained pets recognition & "a photo of a [CLASS], a type of pet." \\
StanfordCars   & 196  & 6,509 & 1,635 & 8,041 & Fine-grained car recognition &"a photo of a [CLASS], a type of flowers." \\
Flowers     & 102  & 4,093 & 1,633 & 2,463 & Fine-grained flowers recognition &"a photo of a [CLASS]."\\
Food101      & 101  & 50,500 & 20,200 & 30,300 & Fine-grained food recognition & "a photo of a [CLASS], a type of food." \\
\midrule
ImageNetV2 &1000 &N/A &N/A &10,000 &Improve ImageNet test reliability & "a photo of a [CLASS]."\\
ImageNet-Sketch &1000 &N/A &N/A &50,899 &Improve ImageNet test reliability & "a photo of a [CLASS]."\\
ImageNet-A &1000 &N/A &N/A &7,500 &Improve ImageNet test reliability & "a photo of a [CLASS]."\\
ImageNet-R &1000 &N/A &N/A &30,000 &Improve ImageNet test reliability & "a photo of a [CLASS]."\\
\bottomrule
\end{tabular}
}
\end{table*}
\begin{figure}[t]
 \centering
 \begin{subfigure}[b]{0.495\linewidth}
  \centering
  \includegraphics[width=\linewidth]{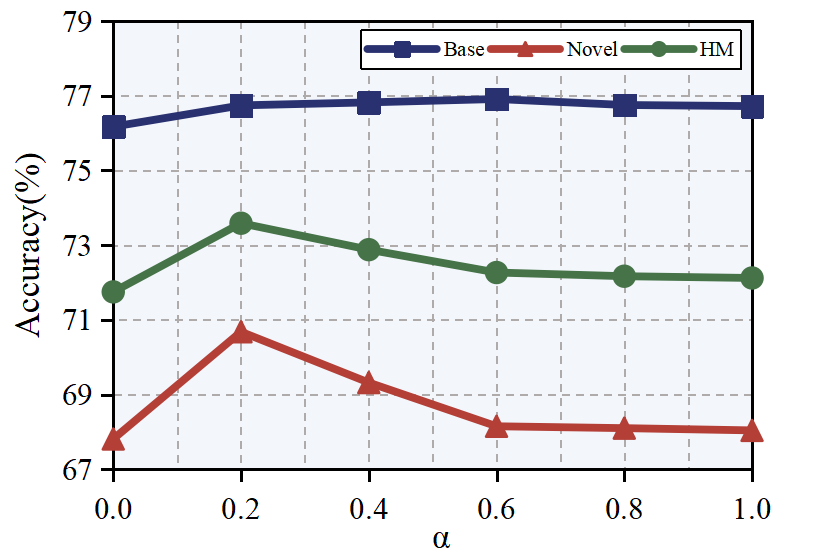}
  \caption{Aggregation Coefficient .}
  \label{local_feat1}
 \end{subfigure}
 \hfill
 \begin{subfigure}[b]{0.493\linewidth}
  \centering
  \includegraphics[width=\linewidth]{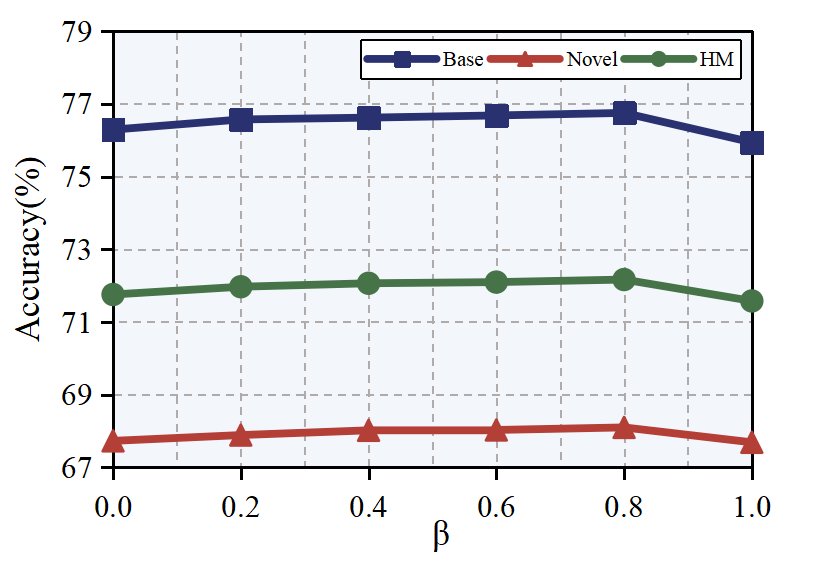}
  \caption{Momentum Coefficient.}
  \label{local_feat2}
 \end{subfigure}
 \vspace{-1em}
 \caption{The impact of different aggregation coefficient and momentum coefficient. }
 \label{local_feat3}
\end{figure}
\begin{table}[h]   
  \centering
  \small
  \caption{Ablation experiments on different backbones.}
  \scalebox{1}{
  \begin{tabular}{c|c|ccc}
    \toprule
     Backbone & Parameters & Base & Novel & HM \\
     \midrule
    ViT-B/16 & 151M & 85.64 & 80.37 & 83.01 \\
    ViT-L/14 & 427M  &85.68 &81.29 & 83.43\\
    \bottomrule
  \end{tabular}}
  \label{tab:backbone}
\end{table}
\begin{table}[h]
\setlength{\tabcolsep}{4pt} 
    \centering
    \small
    \caption{The method chosen for  Image/Text Representative Mapping Module.}
    \begin{tabular}{c|ccc|c}
        \toprule
        Method & Base & Novel & HM &FPS\\
        \hline
        liner+liner & 76.27 &70.98 & 73.53 & 135.19K \\
        trans+trans & 77.64 & 71.75 & 74.58 & 86.89K \\
        original &  77.62 &  71.71 &  74.50 &131.25K \\
        \bottomrule
    \end{tabular}
    \label{tab:results4}
\end{table}


\section{Effects of Coefficient $\alpha$ and $\beta$}
Hyperparameters $\alpha$ and $\beta$ control original information retention and filtered knowledge preservation, respectively. In Fig.\ref{local_feat1}, accuracy on base classes remains stable with increasing $\alpha$, while novel classes peak then decline, suggesting overfitting from fine-grained feature dependence.
Meanwhile, increasing $\beta$ yields gentle rise-then-fall trends for both Base and Novel, confirming the discriminative token selection.

\section{ Hyperparameter Analysis of Optimization Objectives.}
\begin{figure*}[t]
  \centering
	 \subfloat[]{\includegraphics[width=0.3\linewidth]{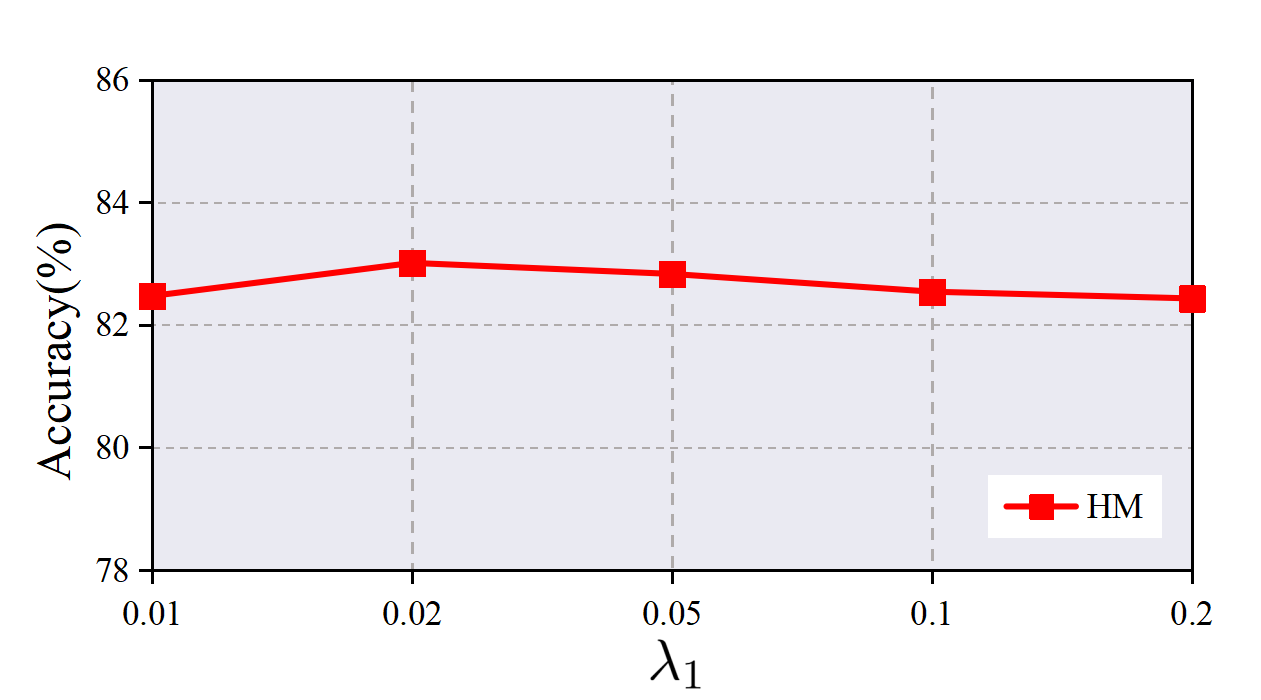}}
  \label{1a}
  \hfill
	 \subfloat[]{\includegraphics[width=0.32\linewidth]{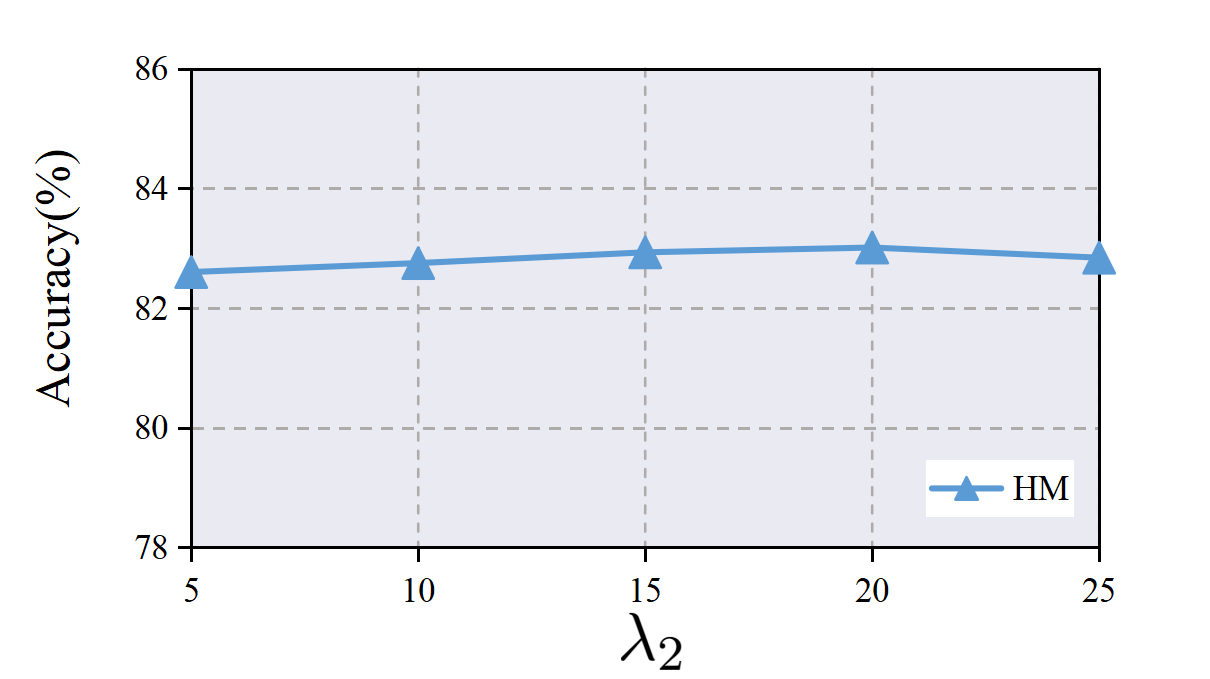}}
  \label{1b}
  \hfill
	 \subfloat[]{\includegraphics[width=0.3\linewidth]{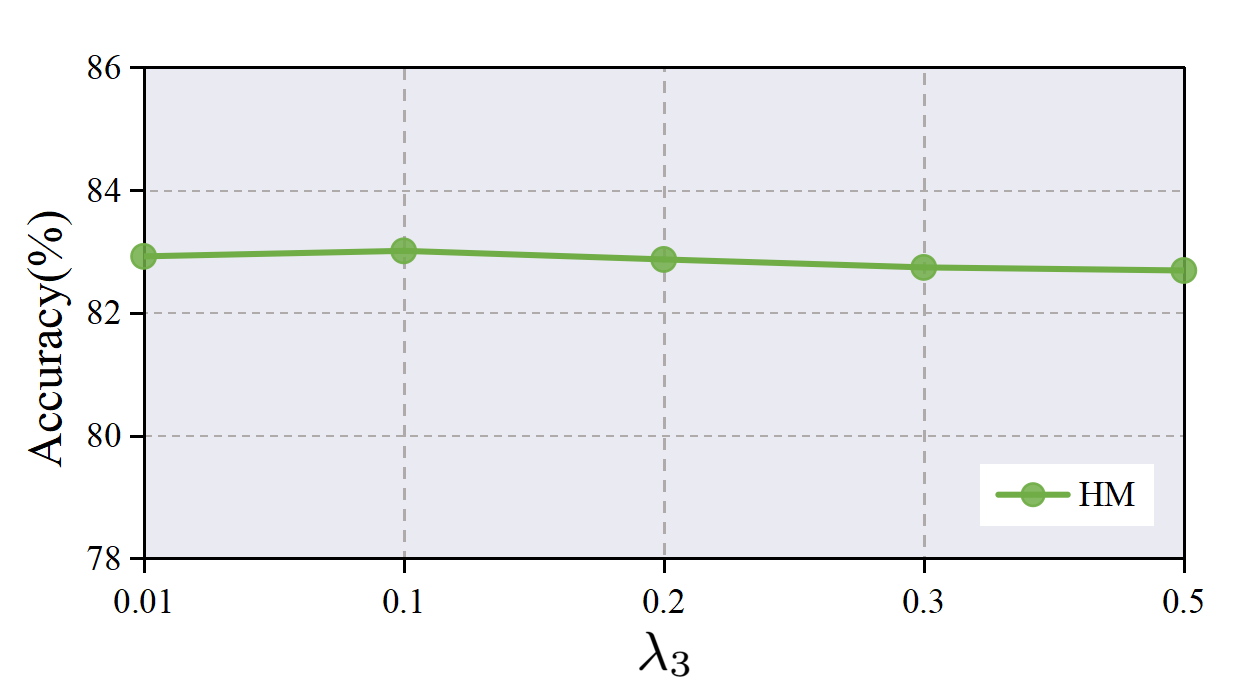}}
	 \caption{The effect of different loss balance parameters $\lambda_{1}$, $\lambda_{2}$, and $\lambda_{3}$ on the model classification accuracy.}
	 \label{senmainloss} 
\end{figure*}

Our systematic investigation of the balancing hyperparameters in Eq.\ref{total} reveals important insights into the method's behavior. Through controlled experiments where we varied individual parameters while fixing others, we observe that the method demonstrates consistent performance across a wide range of configurations, highlighting its robustness and broad applicability to different pre-trained models, shown in Fig.\ref{senmainloss}. However, the performance analysis also identifies critical limitations that tremendous values of $\lambda_{1,2,3}$ lead to noticeable degradation in model performance. This suggests that while the balancing terms are essential for proper alignment, pushing them too far can be counterproductive. The performance drop likely stems from two interrelated factors: first, excessive alignment may force the model to capture artifactual correlations in the training data, leading to overfitting; second, overly strong regularization can constrain the model's capacity to learn meaningful feature representations. These findings emphasize the importance of finding an appropriate balance in parameter settings, where sufficient alignment is achieved without compromising the model's learning capability. The demonstrated robustness across parameter variations further confirms the method's reliability for practical deployment scenarios.

\section{More Few-shot Learning Results}
\vspace{-0.5em}
\begin{figure*}
 \centering
  \includegraphics[width=1\linewidth]{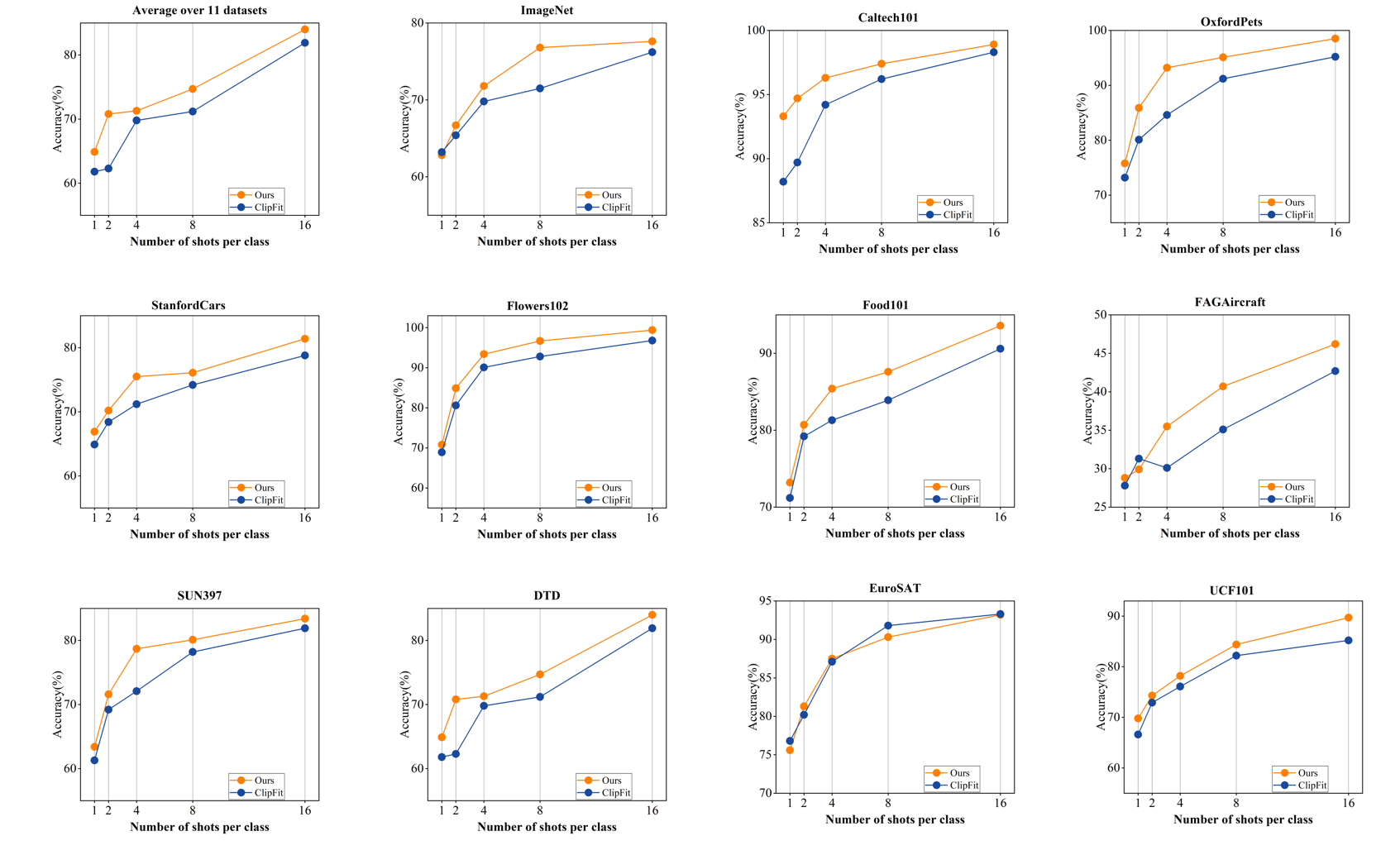}
  \caption{Performance of few-shot learning across 11 datasets compared with CLIPFit~\cite{li2024vision}. The result demonstrates that our method shows better performance than CLIPFit, even with fewer parameters and fewer tokens. }
  \label{fig:TCP2}
\end{figure*}

We adopted the few-shot evaluation protocol from~\cite{radford2021learning}, evaluating our method's ability to acquire task-specific knowledge through 1,2,4,8 and 16-shot learning scenarios while measuring classification accuracy.
In Fig.\ref{fig:TCP2}, we further conducted a visual comparison between our method and CLIPFit, demonstrating superior performance across all 11 datasets.
\section{Effects of Different Backbones of CLIP}
\vspace{-0.5em}
The choice of backbone networks with varying parameter sizes significantly influences model performance. 
To systematically evaluate our method's compatibility with different architectures, we conduct extensive experiments across multiple backbone networks (Table \ref{tab:backbone}). 
The results reveal a consistent trend: model performance scales positively with increasing network capacity, demonstrating our approach's strong adaptability to different architectural scales.
Notably, performance gains exhibit diminishing returns while smaller networks show limited capability due to constrained feature extraction capacity; the performance improvement becomes more pronounced as network size increases.
This pattern suggests that our method effectively leverages the enhanced representational power of larger networks to capture richer feature hierarchies while maintaining stable performance across different architectural scales.

\section{Design of Image/Text Representative Mapping Module.}
\label{map}
We derive the final representative tokens through the Image/Text Mapping Module.
In Table~\ref{tab:results4}, we contrast the methodological designs employed for alignment. We observe that while employing simple linear layers for multimodal processing improves computational efficiency, it leads to noticeable accuracy degradation. Conversely, adopting full transformer architectures yields marginal accuracy gains over current methods while significantly compromising computational efficiency.
This occurs because text tokens inherently encode simpler information compared to visual tokens. Overly complex architectures (e.g., transformers) prove less effective for processing such straightforward patterns, where lightweight linear layers suffice.
\vspace{-1em}
\begin{table}[h]
\setlength{\tabcolsep}{4pt} 
    \centering
    \small
    \caption{Effects of whether to recalculate score.}
    \begin{tabular}{c|ccc}
        \toprule
        Method & Base & Novel & HM \\
        \hline
        \textit{Spotlighter} $w/o$ recaculate & 77.53 & 71.67 &74.48    \\
        \textit{Spotlighter} $w$ recaculate &  77.62 &  71.71 &  74.55   \\
        \bottomrule
    \end{tabular}
    \label{tab:results7}
\end{table}
\section{Effects of whether to recalculate activation score.}
\vspace{-0.5em}
During the secondary classification of activated tokens, we rematch them with prototypes and recompute their activation scores to choose the new top-k. Alternatively, one could directly stratify the top-k activated tokens without recalibration.
The Table~\ref{tab:results7} demonstrates that recalibration yields superior performance compared to direct selection.
This improvement stems from the progressively enriched semantic information encapsulated in the updated prototypes. By rematching and recomputing activated tokens against these refined prototypes, we more accurately identify tokens with the highest semantic density.
\section{Effects of Different Numbers of Activated Tokens.}
\begin{figure}[t]
 \centering
\includegraphics[width=1\linewidth]{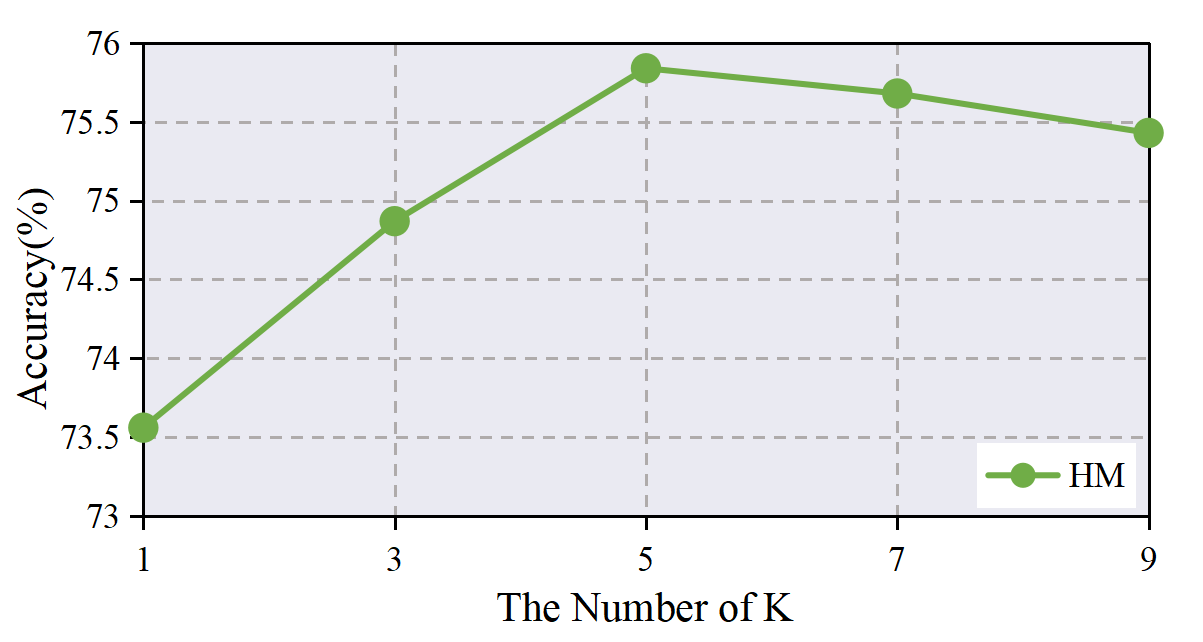}
  \caption{The impact of different activated and representative tokens. }
  \label{top-k}
  \vspace{-1em}
\end{figure}

The token count hyperparameter \textit{k} controls the number of activated and representative tokens.
In Fig.\ref{top-k}, we analyze the impact of different numbers.
The results show that when the number is small, chosen tokens can obtain limited information, but when the number increases, too many tokens decrease speed and obtain noise.
 
\section{Effects of the Chosen Tokens.}
In the semantic memory bank, we select the top-k tokens with the highest activation scores to capture the most representative features of each category. To validate this choice, we conduct additional experiments, shown in Table \ref{tab:results67}. Retaining the bottom-k tokens instead leads to substantially lower performance on ImageNet, indicating that these tokens mainly encode noise. Conversely, removing the top-k tokens also results in inferior accuracy and efficiency despite retaining more tokens, confirming that the most representative features are concentrated in the top-k tokens.
\begin{table}[h]
\setlength{\tabcolsep}{4pt} 
    \centering
    \small
    \caption{Effects of the chosen tokens.}
    \begin{tabular}{c|ccc}
        \toprule
        Method & Base & Novel & HM \\
        \hline
        Retaining the bottom-k & 30.18 & 26.25 &28.26   \\
        Removing the top-k &  62.87 &  53.76 &  57.85   \\
        \rowcolor{blue!10}
         Retaining the top-k  & 77.62 &71.71 &74.55\\
        \bottomrule
    \end{tabular}
    \label{tab:results67}
\end{table}

\end{document}